\documentclass[conference]{IEEEtran}
\IEEEoverridecommandlockouts
\usepackage{cite}
\usepackage{amsmath,amssymb,amsfonts}
\usepackage{algorithmic}
\usepackage{graphicx}
\usepackage{textcomp}
\usepackage{xcolor}
\usepackage[utf8]{inputenc}
\usepackage{tipa}
\usepackage{comment}
\usepackage{xcolor}
\usepackage{booktabs}
\usepackage{multirow}
\usepackage{footnote}
\usepackage{tablefootnote}
\usepackage{makecell}
\usepackage{hyperref}
\usepackage{comment}
\usepackage{enumitem}
\usepackage{adjustbox}
\def\BibTeX{{\rm B\kern-.05em{\sc i\kern-.025em b}\kern-.08em
    T\kern-.1667em\lower.7ex\hbox{E}\kern-.125emX}}
\begin{document}

\title{Study of Indian English Pronunciation Variabilities relative to Received Pronunciation}
\makeatletter
\newcommand{\newlineauthors}{%
  \end{@IEEEauthorhalign}\hfill\mbox{}\par
  \mbox{}\hfill\begin{@IEEEauthorhalign}
}
\makeatother

 \author{\IEEEauthorblockN{Priyanshi Pal}
 \IEEEauthorblockA{\textit{Electrical Department} \\
 \textit{Indian Institute of Science }\\
 Bengaluru 560012, India \\
 priyanshipal53@gmail.com}
 \and
 \IEEEauthorblockN{Shelly Jain}
 \IEEEauthorblockA{\textit{Language Technologies Research Center,} \\
 \textit{IIIT Hyderabad}\\
 Hyderabad, 500032, India \\
 shelly.jain@research.iiit.ac.in}
 
\and
 \IEEEauthorblockN{Anil Vupalla}
 \IEEEauthorblockA{\textit{Language Technologies Research Center,} \\
  \textit{IIIT Hyderabad}\\
 Hyderabad, 500032, India \\
anil.vuppala@iiit.ac.in}

 \newlineauthors
 \IEEEauthorblockN{Chiranjeevi Yarra}
\IEEEauthorblockA{\textit{Language Technologies Research Center,} \\
  \textit{IIIT Hyderabad}\\
 Hyderabad, 500032, India \\
chiranjeevi.yarra@iiit.ac.in}
 \and
\IEEEauthorblockN{Prasanta Kumar Ghosh}
\IEEEauthorblockA{\textit{Electrical Department} \\
 \textit{Indian Institute of Science }\\
 Bengaluru 560012, India \\
prasantg@iisc.ac.in}
 
 }
\maketitle

\begin{abstract}
Analysis of Indian English (IE) pronunciation variabilities are useful in building systems for Automatic Speech Recognition (ASR) and Text-to-Speech (TTS) synthesis in the Indian context. Typically, these pronunciation variabilities have been explored by comparing IE pronunciation with Received Pronunciation (RP). However, to explore these variabilities, it is required to have labelled pronunciation data at the phonetic level, which is scarce for IE. Moreover, versatility of IE stems from the influence of a large diversity of the speakers' mother tongues and demographic region differences. 
Prior linguistic works have characterised features of IE variabilities qualitatively by reporting phonetic rules that represent such variations relative to RP. The qualitative descriptions often lack quantitative descriptors and data-driven analysis of diverse IE pronunciation data to characterise IE on the phonetic level. To address these issues, in this work, we consider a corpus, Indic TIMIT, containing a large set of IE varieties from 80 speakers from various regions of India. 
We present an analysis to obtain the new set of phonetic rules representing IE pronunciation variabilities relative to RP in a data-driven manner. We do this using 15,974 phonetic transcriptions, of which 13,632 were obtained manually in addition to those part of the corpus.
Furthermore, we validate the rules obtained from the analysis against the existing phonetic rules to identify the relevance of the obtained phonetic rules and test the efficacy of Grapheme-to-Phoneme (G2P) conversion developed based on the obtained rules considering Phoneme Error Rate (PER) as the metric for performance.
\end{abstract}

\begin{IEEEkeywords}
Indian English, Pronunciation Analysis, Received pronunciation, Phonetic rules, Rule validation 
\end{IEEEkeywords}

\section{Introduction}
India is a linguistically diverse country having more than $1,369$ mother tongues \cite{chandramouli2011census}. The various languages spoken in India make use of a vast number of vowels and consonants, approximately $18$ and $35$ respectively \cite{kishore2002data}. Indian English (IE) pronunciation is affected by the varying influence of Indian native languages which use many of these vowels and consonants. These variations pose a challenge in automatic speech recognition (ASR) and text-to-speech (TTS) synthesis systems in the Indian context. 
Consequently, these systems are rendered ineffective or yield performance degradation which could be due to the inadequacy of labelled pronunciation data, which is lacking for Indian English speech \cite{sitaram2018discovering}.

For better performance of speech systems on non-native speech, pronunciation models need to consider pronunciation variations influenced by the native language.
\cite{VigneshSRupak} concluded that for the better pronunciation modelling of a language that was non-native to the speaker, the characteristics of the speaker's native language must be considered in the modelling.
Additionally, the differences in the phonemic inventory of various native Indian languages and English (as a second language) play a crucial role in a non-native Indian speaker's pronunciation of English phonemes.
Typically, an Indian speaker is inclined to map English phonemes to the closest phoneme present in their native language \cite{7875936}. 
As suggested in \cite{kumar2007building}, a phoneme set developed to incorporate distinct characteristics of IE phonology can facilitate better pronunciation models for non-native speech.
Approaches such as appropriate selection and optimisation of the phoneme set being considered can increase the effectiveness of speech systems for non-native speech.
\cite{vazhenina2011phoneme} reported that speech recognition was more effective with phoneme set selection techniques for phoneme and word level speech recognition.

Considering these factors, there is a need to study the IE pronunciations at the phonetic level to improve the speech systems for Indian speakers.
Prior studies in the Indian context done to facilitate the adaptability of speech systems for non-native Indian speech are as follows. \cite{anil2016phoneme} reported phoneme selection rules for better naturalness and intelligibility in TTS for Marathi.
\cite{sitaram2018discovering} showed that certain IE accents are more recognisable than others, suggesting their suitability as canonical IE accents. \cite{huang2020construction} developed a linguistically-guided IE pronunciation dictionary for ASR by modifying the North American English (NAE) pronunciations in CMUdict \cite{weide2005carnegie} to IE using observed IE phonological features.
For the few phonemes listed for comparison between NAE and IE pronunciations in IPA, the methodology to obtain phonological feature of IE is unclear due to the fact that ARPAbet is used in CMUdict. Hence, the peculiarities of IE obtained by comparing the canonical NAE pronunciations seem unsuitable. Other works in the Indian context have also studied phonetics and its influences, especially for particular Indian native languages. For instance, \cite{article} examined the phonetics of Telugu speakers' L2 English. 

In prior work, there has been a lack of approaches that focus on analysing sizeable datasets which are diverse in IE pronunciation variabilities, using data-driven means. Typically, this results in capturing very few pronunciation variabilities in IE.
Qualitative observations about various phonetic features of IE can be informative; however, the quantitative metrics used to describe qualitative observations can reveal the prevalence and significance of those observations. Furthermore, the data-driven rules are inherently dependent on the properties of the data being used. In order to study the characteristics of IE using phonetic transcriptions, it must be ensured that latter is reliable, consistent and representative of IE.
This paper addresses these gaps by performing a data-driven analysis of phonetic transcriptions obtained by considering speech recordings in a linguistically diverse, Indic TIMIT corpus \cite{yarra2019indic}. 

We gather existing qualitative phonetic rules relative to RP and report quantitative metrics to represent the prevalence of the phonetic features in IE and their probability of being representative of IE.
We also present new rules found through our data analysis, which have not been discussed in existing literature. Finally, we demonstrate the benefit of the obtained rules in building a Grapheme-to-Phoneme (G2P) conversion system for automatic generation of IE pronunciations.


\section{Data Annotation and Pre-processing}
\label{sec:data}

\subsection{Indic TIMIT Corpus}

We consider the speech data from Indic TIMIT \cite{yarra2019indic} corpus for our work. 
In the corpus, $80$ Indian English L2 speakers were considered from $6$ regions of India, namely -- North-East, East, North, Central, West, and South. From all these regions, 
speakers were recorded while speaking TIMIT stimuli \cite{zue1990speech}, where each speaker was recorded for $2,342$ stimuli. The age of subjects ranged from $18$-$60$ years, and they were either students or staff from the Indian Institute of Science, Bangalore, India.
Cumulatively from all $80$ speakers, a total of $240$ hours of speech data was obtained.
From the considered $6$ regions, a total of $5$ groups were formed, based on regions of the speakers' native language, as described below:

\begin{description}
   \item \textit{Group $1$ (North East and East Regions)}: Maithili, Nepali, Oriya, Bengali, Assamese, Dimasa, Mog, and Manipuri.
   \item \textit{Group $2$ (North and Central Regions)}: Malwi, Marwari, Punjabi and Hindi.
   \item \textit{Group $3$ (West Region)}: Gujarati, Konkani, and Marathi.
   \item \textit{Group $4$ (Upper South Region)}: Kannada and Telugu
   \item \textit{Group $5$ (Lower South Region)}: Malayalam and Tamil.
\end{description}

The languages in these groups were identified based on their originating language families and also influenced by other language families. The languages in Groups $1$, $2$, and $3$ originate from the Indo-Aryan language family, except for Dimasa, Mog and Manipuri, which are Tibeto-Burman languages. Assamese and Nepali are influenced by the Tibeto-Burman language family. Assamese and Bengali are also influenced by Austro-Asiatic language family. The languages in Groups $4$ and $5$ originate from Dravidian language family, wherein languages in the former group are also influenced by Indo-Aryan language family. The considered languages in these groups are spoken in proximate regions. Using information from these groups, further annotation is done.
Since a large majority of the Indian population speak the languages considered in the corpus, subjects from these native languages were considered sufficient to cover the accent variabilities in IE.


\begin{savenotes}
\begin{table*}[ht]
    \centering
    \caption{ \text{Phonetic Rules mentioned in Literature}\vspace*{-1mm}}
    \begin{tabular}{cccc c ccccc}
    \cmidrule(lr){1-4}\cmidrule(lr){6-10}
    \multicolumn{4}{c}{\textbf{General IE Phonetic Rules}} && \multicolumn{5}{c}{\textbf{Native Language Specific Phonetic Rules}} \\
    \cmidrule(lr){1-4}\cmidrule(lr){6-10}
    \textbf{No.} & \textbf{RP} & \textbf{IE} & \textbf{References} && \textbf{No.} & \textbf{RP} & \textbf{IE} & \textbf{Native Language}  & \textbf{References}\\
    \cmidrule(lr){1-4}\cmidrule(lr){6-10}

        1 & /\textipa{E}/ & /\textipa{e}/ or /\textipa{e:}/ & \cite{bansal1994spoken} && 1 & /\textipa{S}/ & /\textipa{s}/ & Hindi, Telugu, Bengali, Bihari & \cite{sailaja2009indian, sirsa2013effects} \\

        2 & /\textipa{2}/ & /\textipa{@}/ & \cite{bansal1994spoken} &&  2 & /\textipa{z}/ & /\textipa{s}/ & Hindi, Telugu, Bengali, Bihari & \cite{sailaja2009indian} \\

        3 & /\textipa{d}/, /\textipa{t}/ & /\textipa{\:d}/, /\textipa{\:t}/ & \cite{bansal1990pronunciation, mesthrie2008introduction, wells1982accents, sailaja2012indian} &&  3 & /\textipa{I}/ & /\textipa{i}/ & Assamese, Bengali, Bihari Hindi, Oriya & \cite{bansal1994spoken} \\
        4 & /\textipa{T}/ & /\textipa{\|[t\super h}/, /\textipa{\|[t}/ & \cite{gargesh2008indian, sailaja2009indian} && 4 & /\textipa{v}/ & /\textipa{bh}/ & Bengali, Oriya, Assamese & \cite{sailaja2009indian} \\ 
    
        5 & /\textipa{D}/ & /\textipa{\|[d}/ & \cite{gargesh2008indian, sailaja2009indian} && 5 & /\textipa{Z}/ & /\textipa{dZ}/ & Kashmiri & \cite{sailaja2009indian} \\
    
        6 & /\textipa{n}/, /\textipa{l}/ & /\textipa{@ n}/, /\textipa{@ l}/ & \cite{bansal1990pronunciation} &&  6 & /\textipa{f}/ & /\textipa{ph}/ & Gujarati, Marathi & \cite{sailaja2009indian} \\
        \cmidrule(lr){1-4}\cmidrule(lr){6-10}
    \end{tabular}
\label{tab:LRT}
\end{table*}
\end{savenotes}
\subsection{Data Annotation}

In Indic TIMIT, two linguists had transcribed a subset of the recordings, numbering $2,342$, containing languages from all $5$ groups. The phonetic transcriptions in an utterance were obtained using an online interface that was presented to the linguists for annotations.

Apart from the subset pre-existing in the Indic TIMIT corpus, we collected annotations for $13,632$ recordings, totalling $15,974$ recordings worth of phonetic transcriptions for the analysis. They were phonetically transcribed sequentially into a total of $5$ groups such that each group covered languages from all $5$ region-based groups. This was done by considering one of the linguists who annotated a subset of transcriptions for Indic TIMIT Corpus. The linguist is affiliated with Spire Lab at the Electrical Engineering Department, Indian Institute of Science.
We believe that the collected phonetic transcriptions could include the phonetic variations resulting from different native languages of the majority of the Indian population.

A total of $108$ IPA symbols were used for transcribing.
The consistency of transcriptions was accessed by calculating Intra-Rater Agreement using Cohen's Kappa Score \cite{cohen1960coefficient}. This agreement score is robust compared to simpler metrics like percentage error since it eliminates the possibility of the agreement occurring by chance. Typically, the Kappa score is calculated between two phonetic transcriptions of the same lengths. If they are unequal due to insertions or deletions of phonemes, they are made equal by inserting dummy symbols.
The consistency was calculated twice, using different approaches.
For the first approach, each group consisted of $200$ transcribed audio files that were repeated. Among the repetitions, the Kappa score was calculated, and this was done for all $5$ groups. The mean Cohen's Kappa Score was $0.827$ for all groups. 
In the second approach, we randomly shuffled transcriptions from all groups of different speakers speaking the same prompt. These shuffled transcriptions were given for correction.
The mean Cohen's Kappa score for this approach was $0.84$.
Generally, scores above $0.80$ indicate strong agreement; therefore, the transcriptions have excellent annotation consistency.

\subsection{Data Pre-Processing}
\label{sec:preproc}

To perform analysis of IE pronunciation, a pronunciation lexicon was created considering the $15,974$ transcriptions from all $5$ groups.
The lexicon contains $16,664$ entries, where each entry contains words and their corresponding pronunciation using IPA notation. Considering the existing literature in which IE pronunciation variations were described relative to RP, we also considered the RP canonical transcriptions obtained using BEEP pronunciation lexicon \cite{robinson1996beep} to compare with IE for the analysis. The phone set of the BEEP lexicon is an extension
of ARPAbet \cite{robinson1995wsjcamo}. It was converted into IPA for comparison with phone-level IPA transcriptions of our speech data. The words which contained ``-" (ex: audio-visual) and were absent in BEEP lexicon were additionally added by considering the pronunciations of individual words which were already available in the lexicon. Most words with ``-" were covered except $\sim$14. Apart from these words, there were $\sim$117 words which were not present in BEEP lexicon, hence those words were not used. The phonetic transcriptions in our lexicon were mapped to that of the RP, for the words existing in our transcriptions and the BEEP lexicon. This resulted in a set of rules indicating the pronunciation variabilities in IE with respect to RP. The analysis based on these rules is presented in Section \ref{sec:analysis}.

\section{Indian English in Linguistic Literature}\vspace*{-0.5mm}
\label{sec:ling_lit}

The influence of Indian native languages on the L2 English of Indian speakers attributes to the characteristic features of Indian English. Few linguistic works discussed these characteristics of IE relative to RP in the past, as mentioned in Table \ref{tab:LRT} and within this section. Considering these, we have assimilated the phonetic rules mentioned in the works. The phonetic rules based on English pronunciation, as spoken by the Indian population regardless of their native language, are considered as \textit{General IE Phonetic Rules} in Table \ref{tab:LRT}. These features have been collectively described in the literature as characteristic identifiers of IE.
It is worth mentioning that although under General IE features, there is a widely discussed /\textipa{v}/-/\textipa{w}/ merger in Indian English; however, recent studies have shown that it is not as strong a feature of IE as it was deemed to be \cite{chand2009v}. Therefore, it is not considered in our assimilated list of general IE rules from literature. The table also includes the phonetic rules which are specific to the native language of a speaker. We also describe the phonetic rules based on context, such as the position of vowels and consonants in a word. These are categorised into context dependent phonetic rules, which are not listed in Table \ref{tab:LRT}.\vspace*{-1.5mm}


\subsection{Context Dependent Phonetic Rules}
\label{sec:ss_ling_lit}

\begin{enumerate}
\item \textbf{Insertion or Omission of Phoneme:}
  In regions like Uttar Pradesh and Bihar, a short vowel /\textipa{I}/ is prefixed at word-initial positions, as the following: ``speech" becomes [\textipa{Ispi:tS}] and ``school" becomes [\textipa{isku:l}] \cite{mesthrie2008introduction, kachru2005asian}. Few speakers add a semivowel before an initial vowel. Some examples would be, ``every" ([\textipa{jevri}]), ``about" ([\textipa{jebau\:t}]), and ``old" ([\textipa{wo:l\:d}]) \cite{wells1982accents}. Conversely, according to \cite{mesthrie2008introduction}, sometimes people also tend to omit the semivowels /\textipa{j}/ and /\textipa{w}/. ``Yet" is realized as [\textipa{E\:t}], ``won't" as [\textipa{o:n\:t}].
\item \textbf{Rhoticity:} In the IE pronunciation of certain words, especially the ones ending with the letter `r', rhoticity is found in the pronunciation \cite{wells1982accents}. For example, as ``letter" ends with /\textipa{r}/, it is realised as [\textipa{@r}]. However, people who are highly fluent in English realise it as the non-rhotic  [\textipa{@}]. /\textipa{r}/ is usually retained after a vowel in IE pronunciation \cite{kachru_1994}. However, whether IE is rhotic or non-rhotic is not unanimously concluded in literature. \cite{sailaja2012indian} mentioned that although non-rhoticity is not governed by region, it is prevalent across regions. 

\item \textbf{Monophthongisation of Diphthongs:} 
A majority of the Indian population uses monophthongs in their English where diphthongs are used in RP \cite{kachru_1994}.
For the diphthongs /\textipa{eI}/ and /\textipa{@U}/, the corresponding monophthongs /\textipa{e:}/ and /\textipa{o:}/ are used \cite{bansal1990pronunciation}. 
In certain contexts, such as word-final positions, these long vowels can be reduced to short vowels. 
For instance, in words like ``today", these vowels are reduced to /\textipa{e}/ and /\textipa{o}/ \cite{sailaja2009indian}. In words similar to ``near" and ``square" where the vowel is succeeded by /\textipa{r}/ (i.e. /rV/), such as ``period" and ``area", IE generally uses /\textipa{i}/ and /\textipa{e}/ instead of /\textipa{I@}/ and /\textipa{e@}/ respectively \cite{wells1982accents}. 

\item \textbf{Word-specific Contexts:}

1) In the ``-ed" inflections which follow voiceless consonants, IE shows a greater use of /\textipa{d}/ over /\textipa{t}/. Some examples include words like ``traced" as [\textipa{tre:sd}] (IE) instead of [\textipa{tre:st}] (RP), and ``packed" as [\textipa{p\ae kd}] (IE) in place of [\textipa{p\ae kt}] (RP) \cite{mesthrie2008introduction}. 

2) Gemination is a common phenomenon, especially in Dravidian languages. Double consonants in written English are often geminated. Few examples are: ``matter" [\textipa{m\ae tt@r}], ``innate" [\textipa{Inne:t}], and ``illegal" [\textipa{Illi:g@l}] \cite{kachru_1994}.\vspace*{-0.5mm}
\end{enumerate}

\section{Data Analysis}\vspace*{-1mm}
\label{sec:analysis}

\subsection{Procedure}
In our data analysis, we aim to observe the variabilities of phonemes used in IE to those in RP. For this, we employ many-to-many (m2m) aligner \cite{jiampojamarn2007:}, which performs alignment followed by classification. Firstly, the phonemes in RP and IE pronunciation are aligned such that one or many phonemes of RP have the corresponding aligned IE phoneme(s) and vice versa. In addition to these alignments, we also obtain confidence score from the m2m aligner indicating the likelihood between each set of aligned phonemes. We consider this confidence value (C.V.) for our analysis.

Typically, m2m aligner is used for the prediction of phonemes, given graphemes. Therefore, the source is graphemes and the target is phonemes. In our analysis, we consider the source as RP phonemes and the target as IE phonemes for various words in the lexicon. We chose the maximum length specification in m2m aligner as 2 for obtaining alignments.
The training of the m2m aligner is
an extension of the forward-backward training of a
one-to-one stochastic transducer presented in \cite{682181}. 
For classification, to determine if prediction should be based on single or multiple phonemes,``chunking" is performed which is based on instance-based learning \cite{aha1991instance}.
The local classification method uses an instance-based learning
technique as a local predictor to generate a set
of IE phoneme candidates for each RP phoneme chunk, given
its context in the sequence of phonemes denoting the pronunciation. In this process, it produces C.V. for each of aligned set of phonemes.

Since C.V. indicates the likelihood of the IE phoneme(s) for a corresponding RP phoneme(s), we consider these values to validate the rules (phoneme mappings between IE and RP) obtained from the analysis based on aligned set of phonemes with the existing rules reported in the literature.
We also consider the normalised frequency of occurrence corresponding to that rule, to indicate how recurrently it is observed. The frequency of occurrence of a rule is normalised by the total number of occurrences of RP phonemes of that rule.

\begin{table*}[ht]
    \centering
    \caption{Indian English (IE) Phonetic Rules relative to Received Pronunciation (RP).  \textbf{Category 1}: Phonetic Rules mentioned in Literature and observed in Dataset, \textbf{Category 2}: Phonetic Rules observed in Dataset but not discussed in Literature, \textbf{Category 3}: Phonetic Rules mentioned in literature but not observed in the Dataset. `*' Indicates Native language specific rules.}\vspace*{-1mm}
    \begin{adjustbox}{max width=\textwidth}
        \begin{tabular}{ccccc c ccccc}
            \multicolumn{5}{c}{\textbf{Category 1}} && \multicolumn{5}{c}{\textbf{Category 2}}\\
            \cmidrule(lr){1-5}\cmidrule(lr){7-11}
            \textbf{No.} & \textbf{RP} & \textbf{IE} & \textbf{C.V.} & \textbf{Normalised Freq.} && \textbf{No.} & \textbf{RP} & \textbf{IE} & \textbf{C.V.} & \textbf{Normalised Freq.} \\
            \cmidrule(lr){1-5}\cmidrule(lr){7-11}
            1 & /\textipa{E}/ & /\textipa{e}/ & 0.917 & 0.912  &&  1 & /\textipa{U}/ & /\textipa{u}/ & 0.980 & 0.747  \\
            2 & /\textipa{2}/ & /\textipa{@}/ & 0.94 & 0.932  && 2 & /\textipa{aU}/ & /\textipa{au}/ & 0.576 & 0.569  \\
            3 & /\textipa{d}/, /\textipa{t}/ & /\textipa{\:d}/, /\textipa{\:t}/ & 0.964, 0.964 & 0.820, 0.851 && 3 & /\textipa{j U}/ & /\textipa{u}/ & 0.765 & 0.835  \\
            4 & /\textipa{T}/ & /\textipa{\|[t}h/, /\textipa{\|[t}/ & 0.502, 0.45 & 0.453, 0.381 && 4 & /\textipa{\textrhookrevepsilon}/ & /\textipa{@ r}/ & 0.866 & 0.525   \\ 
            5 & /\textipa{D}/ & /\textipa{\|[d}/ & 0.737 & 0.669 && 5 & /\textipa{A}/ & /\textipa{a r}/ & 0.624 & 0.237 \\
            6 & /\textipa{l}/ & /\textipa{@ l}/ & 0.159 & 0.183 && 6 & /\textipa{I d}/ & /\textipa{e \:d}/ & 0.912 & 0.373  \\
            7 & */\textipa{z}/ & /\textipa{s}/ & 0.607(t), 0.557(h), 0.537(b) & 0.584(t), 0.552(h), 0.592(b) &&  7 & /\textipa{S n}/ & /\textipa{@ n}/ & 0.843 & 0.893  \\
            8 & */\textipa{I}/ & /\textipa{i}/ & 0.837 & 0.818 && 8 & /\textipa{@ n}/ & /\textipa{e n}/ & 0.729 & 0.451 \\ 
            \cmidrule(lr){1-5}\cmidrule(lr){7-11}

        \end{tabular}
        \end{adjustbox}
    \vspace{1em}
    \centering
    \begin{adjustbox}{max width=\textwidth}
        \begin{tabular}{cccccc}
            \multicolumn{6}{c}{\textbf{Category 3}}\\
            \toprule
            \textbf{No.} & \textbf{RP} & \textbf{Exp. IE} & \textbf{Obs. IE} & \textbf{C.V. (Obs. IE)} & \textbf{Norm. Freq. (Obs. IE)} \\
            \midrule
            1 & /\textipa{n}/ & /\textipa{@ n}/ & /\textipa{n}/ &  0.873 & 0.902 \\
            2 & */\textipa{S}/ & /\textipa{s}/ & /\textipa{S}/ & 0.402(t), 0.336(h), 0.47(b) & 0.375(t), 0.334(h), 0.420(b) \\
            3 & */\textipa{S}/ & /\textipa{s}/ & /\textipa{S @}/ & 0.508(t), 0.481(h), 0.461(b) & 0.416(t), 0.399(h), 0.389(b) \\
            4 & */\textipa{v}/ & /\textipa{b h}/ & /\textipa{v}/ & 0.964 & 0.942 \\
            5 & */\textipa{f}/ & /\textipa{p h}/ & /\textipa{f}/ & 0.984 & 0.984 \\
            6 & /\textipa{oU}/ & /\textipa{o:}/ & /\textipa{o}/ & 0.925 & 0.731 \\ 
            7 & /\textipa{eI}/ & /\textipa{e:}/ & /\textipa{e}/ & 0.953 & 0.727 \\
            8 & /\textipa{6}/ & /\textipa{O:}/ & /\textipa{O}/ & 0.871 & 0.654\\ 
            9 & \textcolor{red}{/\textipa{@U}/, /\textipa{E@}/, /\textipa{Ie}/, /\textipa{A:}/, /\textipa{O:}/} & - & - & - & - \\ 
            \bottomrule\vspace*{-5.5mm}
        \end{tabular}
        \end{adjustbox}
    \label{tab:all_rules}
\end{table*}
The C.V. and normalised frequency both range from 0 to 1. 
The rules with C.V. of 0.10 and above are considered in this analysis. 
Furthermore, a minimum frequency of occurrence of 150 is also ensured for each rule to avoid C.V. and normalised frequency values derived from low frequency of occurrence of the rule in the data. We grouped the phonetic rules into three categories in Table \ref{tab:all_rules} based on their occurrence in literature, dataset, and as found using data-driven method:\vspace*{-0.5mm}

\begin{itemize}
\item {\textbf{Category 1} - \emph{Phonetic rules mentioned in literature and observed in the dataset.}}

This contains IE phonetic rules which were validated by using data-driven methods. The last two phonetic rules (indicated with asterisk) only apply to a population with specific native languages. 

\item {\textbf{Category 2} - \emph{Phonetic rules observed in dataset, but not discussed in literature}}

This consists of phonetic rules which were observed with high confidence value and normalised frequency. However, discussion regarding them was not found in the linguistic works we studied.

\item {\textbf{Category 3} - \emph{Phonetic rules mentioned in literature but not observed in dataset}}

The phonemes in the last row are not analysed for their related Phonetic rules since they are not present in the canonical transcriptions obtained using phone set in BEEP lexicon. Remaining phonetic rules have been discussed in the literature, however were not prominently observed in our dataset. Either they were not present in our dataset or were not prominent enough to cross our thresholds for confidence value and frequency of occurrence. We also report the phoneme observed (Obs. IE) in our data in place of the expected phoneme (Exp. IE) which is mentioned in the literature.
The rules in rows 1, 6, 7, and 8 correspond to general IE and the rest of the rules are native language specific.
\end{itemize}

\subsection{Discussion}

The three categories of phonetic rules in Table \ref{tab:all_rules} are discussed, considering C.V. and the normalised frequency of occurrence (Normalised  Freq.) from the data. 

\subsubsection{Category 1}

The rules in rows 1-6  correspond to general IE features mentioned in literature, were prominent in our dataset, indicated by high C.V. and Normalised Freq. values. 

For the rule in row 6, although for the phoneme /\textipa{l}/ in R.P, the most commonly observed corresponding phoneme in I.E is /\textipa{l}/ in our data, there is a significant presence of usage of /\textipa{@ l}/ as well. Therefore, this sort of insertion of phoneme happens only sometimes, as mentioned in first point of Section \ref{sec:ss_ling_lit}.

The rules in rows 7 and 8 correspond to specific native languages. The transcriptions obtained from the native speakers of those languages are considered. For row 7, languages are Hindi and Bihari (h), Telugu (t) and Bengali (b). For row 8, Hindi, Bengali, Assamese and Oriya native languages are applicable.

\subsubsection{Category 2}
The phonetic rule in row 1 has not been discussed in literature where comparisons between the RP /\textipa{U}/ and /\textipa{u}/ have been made. However, in the work by \cite{sirsa2013effects}, the existence of a /\textipa{U}/-/\textipa{u}/ merger has been pointed out in general for Gujarati English and Indian English.

Consequently, it is also possible that as a result of the phonetic rule in row 1, 
the rule in row 2 can be observed wherein, for the diphthong /\textipa{aU}/, /\textipa{au}/ is observed instead. 
There might also have been diphthongs where this replacement could be seen; however, such phonetic rules would not have met either the confidence value or minimum frequency of occurrence criteria in our dataset.
Additionally, its influence can also be seen in the rule of row 3. However, rule in row 3 also suggests the deletion of a phoneme. For example, it was mentioned in Section \ref{sec:ling_lit} that in certain contexts, semivowels /\textipa{j}/ and /\textipa{w}/ are omitted. 
Therefore, further investigations regarding context of usage could be useful in understanding the presence of this rule. The rules in rows 4 and 5 could suggest mild rhoticity in the speakers' accents, as mentioned in point 4 of Section \ref{sec:ling_lit}.
In the rule corresponding to row 6, the usage of retroflex /\textipa{\:d}/ is clear from the validation of rule in Category 1, row 3. However, there is little information regarding the presence of /\textipa{e}/ in /\textipa{e \:d}/ or /\textipa{I}/ in /\textipa{I d}/.

Syllabification of /\textipa{n}/ and /\textipa{l}/ as /\textipa{@ n}/ and /\textipa{@ l}/ is discussed in \cite{bansal1990pronunciation}. The presence of rule in row 7 could indicate phoneme insertion, particularly in words ending with ``-tion". For example, in the word ``absorption", the RP pronunciation is [\textipa{@bsOp\textbf{Sn}}], whereas [\textipa{@bsOp\textbf{S@n}}] is the IE alternative. Therefore, presence of RP and IE phonemes in such word contexts could relate to the presence of this rule. This may also be associated to the discussion in \cite{sailaja2009indian}, where the insertion of /\textipa{@}/ in word-final cluster like "lm" in words such as "film" i.e  /\textipa{fIl@m}/ is mentioned. In order to conclusively understand these phonetic rules, analysis of contexts along with native language is needed for the rules in rows 6, 7 and 8.

\subsubsection{Category 3}

Row 1 follows the description of syllabification in Section \ref{sec:ling_lit}. However, unlike the schwa insertion in /\textipa{@ l}/ for /\textipa{l}, presence of /\textipa{@ n}/ wasn't observed for /\textipa{n}/. Instead, the prevalent usage was closer to RP phoneme /\textipa{n}/. 

The phonetic rules in rows 2, 3, 4 and 5 are native language specific. The rules in rows 2 and 3 share the same native languages as row 7 in Category 1. Row 4 is applicable to Bengali, Oriya and Assamese speakers and row 5 for Gujarati or Marathi speakers. These are mentioned in rows 1, 4 and 6 in Table \ref{tab:LRT}.
For all three, we observed the phonemes in IE for the corresponding native languages to be same as RP.
In row 2, apart from the occurrences where the IE phoneme /\textipa{S}/ is the same as RP, we also observed /\textipa{S @}/ in IE, which is listed in row 3. This could indicate the presence of /\textipa{@}/ phoneme insertion.


In row 4, the prominent usage of the phoneme /\textipa{v}/ instead of /\textipa{b h}/ indicates a possibly vanishing /\textipa{v}/-/\textipa{b h}/ substitution, which may be similar to the vanishing /\textipa{v}/-/\textipa{w}/ merger. Similarly, row 5  indicates that /\textipa{f}/ was retained in its original form.

The rules in rows 6, 7 and 8 correspond to the diphthongs in RP often substituted as monophthongs in IE. Contrary to the phonemes being substituted by a long vowel, we observed a wide usage of short vowels, with high prominence. However, in contexts such as the ones which are mentioned towards the end of point 3 in Section \ref{sec:ss_ling_lit},
IE often has short vowels substituting diphthongs. 

The rules for phonemes mentioned in row 9 are absent in the RP canonical transcriptions. Description for some of the rules related to them are as follows. In \cite{bansal1994spoken}, the rule consists of RP phoneme /\textipa{A:}/ and its corresponding IE phoneme /\textipa{a:}/. Additionally, IE /\textipa{6:}/ is mentioned for RP /\textipa{O:}/ in another rule.

Finally, in addition to the rules mentioned in this category, with reference to point 5 under native specific language features in Table \ref{tab:LRT}, the phonetic rule specified for Kashmiri native language is not analysed due to its absence in the languages considered in Indic TIMIT corpus.

\subsubsection{Context Dependent Phonetic Rules}

For the words ending with ``-ed", the usage of IE /\textipa{d}/ instead of RP /\textipa{t}/ was barely observable. 
In our RP pronunciations for ``-ed" ending words, instead of /\textipa{t}/, we observed /\textipa{d}/. This is contradictory to description under point 1 in Word specific Contexts. Furthermore, the IE phoneme /\textipa{\:d}/ for those words was observed, in place of RP phoneme /\textipa{d}/, which is expected.

For gemination, we considered words with consonants such as ``ll", ``nn" and ``tt". Very few instances of gemination by Indian speakers were observed. 
Many Indian languages have gemination in their verbal and orthographic forms, which explains the expectation for a native Indian Language speaker to influence their L2 English similarly. However, a possible explanation for the absence of this behavior in our data could be that in the limited words where context was applicable, speakers pronounced correct phonetic sequence. 
Lastly, we consider the insertion rule corresponding to /\textipa{I}/ insertion as mentioned in point 1, particularly for words starting with ``s". We consider Hindi speakers to validate /\textipa{I}/ insertion. There were very rare instances where this was observed to happen. Apart from this, when word-initial positions were considered for semi-vowel insertion, the occurrences were very few. 

\subsection{Testing Efficacy of G2P system based on Phonetic Rules}

\begin{table}[ht!] \vspace*{-0.25cm}
\centering
\caption{Phoneme Error Rate (PER) for the lexicons}\vspace*{-0.25cm}
\begin{tabular}{cccc}
\cmidrule(lr){1-4}
\textbf{Lexicon} & IE & RP & IE\_PRAG\\
\textbf{PER} & 7\% & 47\% & 25\% \\

\bottomrule 
\end{tabular}

\label{tab:lex_per_table}
\end{table}

We consider Sequitur G2P conversion system \cite{BISANI2008434} to show the effectiveness of the phonetic rules obtained from the proposed analysis. For the experimentation, we consider three pronunciation lexicons. The first one is referred as IE lexicon, which is described in \ref{sec:preproc}. It is constructed using unique pairs of words in the stimuli and their respective annotated phonetic transcriptions. The second lexicon is referred to as RP lexicon, which is BEEP pronunciation lexicon. Finally, the third one is referred to as IE\_PRAG (\textbf{P}honetic \textbf{R}ule based \textbf{A}utomatically \textbf{G}enerated) lexicon, constructed from rules in \ref{tab:all_rules} by substituting the phonemes of RP column in all the pronunciation sequences in the RP lexicon with the phonemes of IE column. Each substitution rule is applied to the fraction (equal to the Normalised frequency in the table) of all possible candidates in the RP lexicon for the rule, chosen randomly. It is observed that the unique words vary in IE\_PRAG, RP and IE lexicon. Thus, a similar approach mentioned in \ref{sec:preproc} is used to consider unique words common across all three. These are found to be a total of $6,720$ out of which the pronunciation entries correspond to $5,376$ (randomly chosen) words from all the three lexicons for training G2P system and the entries of the remaining words for testing. We consider Phoneme Error Rate (PER) as the metric for the evaluation on the test set. From the PER reported in \ref{tab:lex_per_table} with all three lexicons, it is observed that the PER with IE\_PRAG lexicon is lesser than that of the RP lexicon. This shows the benefit of the phonetic rules obtained from the proposed analysis for building lexicon for IE automatically with G2P. Hence, IE\_PRAG lexicon could be helpful in building better ASR and TTS in the Indian context.

\section{Conclusion}\vspace*{-0.5mm}
\label{sec:conc}

Addressing the need to study and analyse IE pronunciation, we used a data-driven approach to explore the pronunciation variabilities of IE relative to RP. For this, we phonetically transcribed $13,632$ utterances taken from the Indic TIMIT speech corpus. Considering a total of $15,974$ phonetic transcriptions, we presented a methodology to extract phonetic rules and validate them for their relevance and significance in the Indian context. In the validation process, we verified the relevance of extracted rules from the data with the rules reported in the existing literature regarding both general and native language specific pronunciation variabilities, along with a set of new and absent rules with respect to the findings reported in the literature. We believed that the indicative rules helped determine relevant IE phonetic tendencies with higher confidence. Furthermore, we compared performance of G2P conversion using lexicons constructed with and without the phonetic rules obtained in the proposed analysis.
Potential applications of these rules include phone set optimisation for IE and modification of pronunciation lexicon for better ASR and TTS performance. Further investigation is needed to analyse the quality of the new set of rules based on the influences from the native language specific patterns. Future directions include identifying the reasons for the absent rules reported in the analysis as well as further investigating the performance changes in ASR or TTS systems using the reported rules. \vspace*{-1.5mm}

\bibliographystyle{IEEEtran}
\bibliography{IEEEabrv,fin_ref}

\begin{thebibliography}{10}
\providecommand{\url}[1]{#1}
\csname url@samestyle\endcsname
\providecommand{\newblock}{\relax}
\providecommand{\bibinfo}[2]{#2}
\providecommand{\BIBentrySTDinterwordspacing}{\spaceskip=0pt\relax}
\providecommand{\BIBentryALTinterwordstretchfactor}{4}
\providecommand{\BIBentryALTinterwordspacing}{\spaceskip=\fontdimen2\font plus
\BIBentryALTinterwordstretchfactor\fontdimen3\font minus
  \fontdimen4\font\relax}
\providecommand{\BIBforeignlanguage}[2]{{%
\expandafter\ifx\csname l@#1\endcsname\relax
\typeout{** WARNING: IEEEtran.bst: No hyphenation pattern has been}%
\typeout{** loaded for the language `#1'. Using the pattern for}%
\typeout{** the default language instead.}%
\else
\language=\csname l@#1\endcsname
\fi
#2}}
\providecommand{\BIBdecl}{\relax}
\BIBdecl

\bibitem{chandramouli2011census}
C.~Chandramouli and R.~General, ``Census of india 2011,'' \emph{Provisional
  Population Totals. New Delhi: Government of India}, pp. 409--413, 2011.

\bibitem{kishore2002data}
S.~Kishore, R.~Kumar, and R.~Sangal, ``A data driven synthesis approach for
  indian languages using syllable as basic unit,'' in \emph{Proceedings of
  Intl. Conf. on NLP (ICON)}, 2002, pp. 311--316.

\bibitem{sitaram2018discovering}
S.~Sitaram, V.~Manjunatha, V.~Bharadwaj, M.~Choudhury, K.~Bali, and M.~Tjalve,
  ``Discovering canonical indian english accents: a crowdsourcing-based
  approach,'' in \emph{Proceedings of the Eleventh International Conference on
  Language Resources and Evaluation (LREC 2018)}, 2018.

\bibitem{VigneshSRupak}
S.~R. Vignesh, S.~A. Shanmugam, and H.~A. Murthy, ``Significance of
  pseudo-syllables in building better acoustic models for indian english tts,''
  in \emph{2016 IEEE International Conference on Acoustics, Speech and Signal
  Processing (ICASSP)}, 2016, pp. 5620--5624.

\bibitem{7875936}
R.~Saikia and S.~R. Singh, ``Generating manipuri english pronunciation
  dictionary using sequence labelling problem,'' in \emph{2016 International
  Conference on Asian Language Processing (IALP)}, 2016, pp. 67--70.

\bibitem{kumar2007building}
R.~Kumar, R.~Gangadharaiah, S.~Rao, K.~Prahallad, C.~P. Ros{\'e}, and A.~W.
  Black, ``Building a better indian english voice using" more data".'' in
  \emph{SSW}.\hskip 1em plus 0.5em minus 0.4em\relax Citeseer, 2007, pp.
  90--94.

\bibitem{vazhenina2011phoneme}
D.~Vazhenina and K.~Markov, ``Phoneme set selection for russian speech
  recognition,'' in \emph{2011 7th International Conference on Natural Language
  Processing and Knowledge Engineering}.\hskip 1em plus 0.5em minus 0.4em\relax
  IEEE, 2011, pp. 475--478.

\bibitem{anil2016phoneme}
M.~C. Anil and S.~Shirbahadurkar, ``Phoneme selection rules for marathi text to
  speech synthesis with anuswar places,'' in \emph{Advances in Signal
  Processing and Intelligent Recognition Systems}.\hskip 1em plus 0.5em minus
  0.4em\relax Springer, 2016, pp. 501--509.

\bibitem{huang2020construction}
X.~Huang, X.~Jin, Q.~Li, and K.~Zhang, ``On construction of the asr-oriented
  indian english pronunciation dictionary,'' in \emph{Proceedings of the 12th
  Language Resources and Evaluation Conference}, 2020, pp. 6593--6598.

\bibitem{weide2005carnegie}
R.~Weide, ``The carnegie mellon pronouncing dictionary [cmudict. 0.6],''
  \emph{Version 0.6. Available at [www. speech. cs. cmu. edu/cgi-bin/cmudict]},
  2005.

\bibitem{article}
K.~Pisegna and V.~Volenec, ``Phonology and phonetics of l2 telugu english,''
  \emph{Studies in Linguistics and Literature}, vol.~5, pp. 46--69, 02 2021.

\bibitem{yarra2019indic}
C.~Yarra, R.~Aggarwal, A.~Rajpal, and P.~K. Ghosh, ``Indic timit and indic
  english lexicon: A speech database of indian speakers using timit stimuli and
  a lexicon from their mispronunciations,'' in \emph{2019 22nd Conference of
  the Oriental COCOSDA International Committee for the Co-ordination and
  Standardisation of Speech Databases and Assessment Techniques
  (O-COCOSDA)}.\hskip 1em plus 0.5em minus 0.4em\relax IEEE, 2019, pp. 1--6.

\bibitem{zue1990speech}
V.~Zue, S.~Seneff, and J.~Glass, ``Speech database development at mit: Timit
  and beyond,'' \emph{Speech communication}, vol.~9, no.~4, pp. 351--356, 1990.

\bibitem{bansal1994spoken}
R.~K. Bansal and J.~B. Harrison, \emph{Spoken English: A Manual of Speech and
  Phonetics}.\hskip 1em plus 0.5em minus 0.4em\relax Sangam, 1994.

\bibitem{sailaja2009indian}
S.~Pingali, \emph{Indian English}.\hskip 1em plus 0.5em minus 0.4em\relax
  Edinburgh University Press, 2009.

\bibitem{sirsa2013effects}
H.~Sirsa and M.~A. Redford, ``The effects of native language on indian english
  sounds and timing patterns,'' \emph{Journal of phonetics}, vol.~41, no.~6,
  pp. 393--406, 2013.

\bibitem{bansal1990pronunciation}
R.~K. Bansal, ``The pronunciation of english in india,'' \emph{Studies in the
  pronunciation of English: A commemorative volume in honour of AC Gimson}, pp.
  219--230, 1990.

\bibitem{mesthrie2008introduction}
R.~Mesthrie, ``Introduction: varieties of english in africa andsouth and
  southeast asia,'' in \emph{4 Africa, South and Southeast Asia}.\hskip 1em
  plus 0.5em minus 0.4em\relax De Gruyter Mouton, 2008, pp. 23--34.

\bibitem{wells1982accents}
J.~C. Wells, \emph{Accents of English: Volume 3: Beyond the British
  Isles}.\hskip 1em plus 0.5em minus 0.4em\relax Cambridge University Press,
  1982, vol.~3.

\bibitem{sailaja2012indian}
P.~Sailaja, ``Indian english: Features and sociolinguistic aspects,''
  \emph{Language and Linguistics Compass}, vol.~6, no.~6, pp. 359--370, 2012.

\bibitem{gargesh2008indian}
R.~Gargesh, ``Indian english: Phonology,'' in \emph{A handbook of varieties of
  English}.\hskip 1em plus 0.5em minus 0.4em\relax De Gruyter Mouton, 2008, pp.
  992--1002.

\bibitem{cohen1960coefficient}
J.~Cohen, ``A coefficient of agreement for nominal scales,'' \emph{Educational
  and psychological measurement}, vol.~20, no.~1, pp. 37--46, 1960.

\bibitem{robinson1996beep}
T.~Robinson, ``Beep dictionary,'' \emph{BEEP dictionary}, 1996.

\bibitem{robinson1995wsjcamo}
T.~Robinson, J.~Fransen, D.~Pye, J.~Foote, and S.~Renals, ``Wsjcamo: a british
  english speech corpus for large vocabulary continuous speech recognition,''
  in \emph{1995 International Conference on Acoustics, Speech, and Signal
  Processing}, vol.~1.\hskip 1em plus 0.5em minus 0.4em\relax IEEE, 1995, pp.
  81--84.

\bibitem{chand2009v}
V.~Chand, ``[v] at is going on? local and global ideologies about indian
  english,'' \emph{Language in Society}, vol.~38, no.~4, pp. 393--419, 2009.

\bibitem{kachru2005asian}
B.~B. Kachru, \emph{Asian Englishes: beyond the canon}.\hskip 1em plus 0.5em
  minus 0.4em\relax Hong Kong University Press, 2005, vol.~1.

\bibitem{kachru_1994}
------, \emph{English in South Asia}, ser. The Cambridge History of the English
  Language.\hskip 1em plus 0.5em minus 0.4em\relax Cambridge University Press,
  1994, vol.~5, p. 497–553.

\bibitem{jiampojamarn2007:}
\BIBentryALTinterwordspacing
S.~Jiampojamarn, G.~Kondrak, and T.~Sherif, ``Applying many-to-many alignments
  and hidden markov models to letter-to-phoneme conversion,'' in \emph{Human
  Language Technologies 2007: The Conference of the North American Chapter of
  the Association for Computational Linguistics; Proceedings of the Main
  Conference}.\hskip 1em plus 0.5em minus 0.4em\relax Rochester, New York:
  Association for Computational Linguistics, April 2007, pp. 372--379.
  [Online]. Available: \url{http://www.aclweb.org/anthology/N/N07/N07-1047}
\BIBentrySTDinterwordspacing

\bibitem{682181}
E.~Ristad and P.~Yianilos, ``Learning string-edit distance,'' \emph{IEEE
  Transactions on Pattern Analysis and Machine Intelligence}, vol.~20, no.~5,
  pp. 522--532, 1998.

\bibitem{aha1991instance}
D.~Aha, D.~Kibler, and M.~Albert, ``Instance-based learning techniques,''
  \emph{Machine Learning}, vol.~6, pp. 37--66, 1991.

\bibitem{BISANI2008434}
\BIBentryALTinterwordspacing
M.~Bisani and H.~Ney, ``Joint-sequence models for grapheme-to-phoneme
  conversion,'' \emph{Speech Communication}, vol.~50, no.~5, pp. 434--451,
  2008. [Online]. Available:
  \url{https://www.sciencedirect.com/science/article/pii/S0167639308000046}
\BIBentrySTDinterwordspacing

\end{thebibliography}
\vspace{12pt}

\end{document}